\documentclass[runningheads]{llncs}
\usepackage[T1]{fontenc}
\usepackage{amsmath}
\usepackage{bbm}
\usepackage{booktabs}  
\usepackage{amsmath, amssymb}
\usepackage{algorithm}
\usepackage{algpseudocode}
\usepackage{graphicx}
\usepackage[htt]{hyphenat}
\usepackage{multirow}
\usepackage{graphicx,verbatim}
\usepackage{color}
\usepackage{xcolor}
\usepackage{caption}
\usepackage{float}
\usepackage{arydshln}
\usepackage{hyperref}
\usepackage{color}

\begin{document}
\title{RL4Med-DDPO: Reinforcement Learning for Controlled Guidance Towards Diverse Medical Image Generation using Vision-Language Foundation Models}

\author{Parham Saremi$^*$
\inst{1,2} \and
Amar Kumar$^*$
\inst{1,2}\and
Mohamed Mohamed\inst{1,2} \and Zahra TehraniNasab\inst{1,2} \and Tal Arbel\inst{1,2}}

\authorrunning{P. Saremi et al.}
\titlerunning{RL4Med-DDPO}

\institute{Center for Intelligent Machines, McGill University, Montreal, Canada \and
Mila - Quebec AI institute, Montreal, Canada \\
\email{parham.saremi@mail.mcgill.ca}\\
    $^*$
equal contribution }

\maketitle              
\begin{abstract}

Vision-Language Foundation Models (VLFM) have shown a tremendous increase in performance in terms of generating high-resolution, photorealistic natural images. While VLFMs show a rich understanding of semantic content across modalities, they often struggle with fine-grained alignment tasks that require precise correspondence between image regions and textual descriptions, a limitation in medical imaging, where accurate localization and detection of clinical features are essential for diagnosis and analysis. To address this issue, we propose a multi-stage architecture where a pre-trained VLFM (e.g. Stable Diffusion) provides a cursory semantic understanding, while a reinforcement learning (RL) algorithm refines the alignment through an iterative process that optimizes for understanding semantic context. The reward signal is designed to align the semantic information of the text with synthesized images. Experiments on the public ISIC2019 skin lesion dataset demonstrate that the proposed method improves (a) the quality of the generated images, and (b) the alignment with the text prompt over the original fine-tuned Stable Diffusion baseline. 
We also show that the synthesized samples could be used to improve disease classifier performance for underrepresented subgroups through augmentation. Our code is accessible through the project website.\footnote{\url{https://parhamsaremi.github.io/rl4med-ddpo}}

\end{abstract}
\section{Introduction}
\begin{figure}[t]
    \centering
    \includegraphics[width=0.85\linewidth]{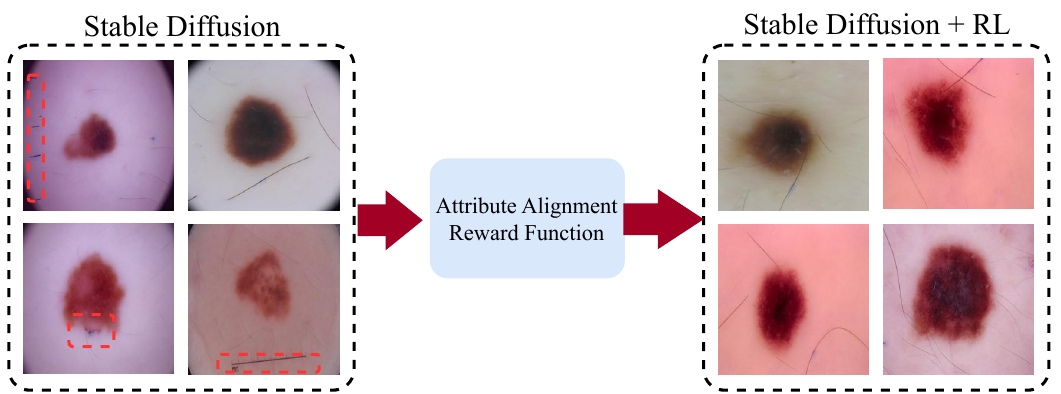}
    \caption{Comparison of synthetic samples generated from Stable Diffusion (left image) and Stable Diffusion with Reinforcement Learning (right image). The text prompt for these image samples was - \texttt{A dermatoscopic image with melanoma showing hairs}. Note the unwanted but relevant \textcolor[HTML]{FF3333}{artifacts} that do not align with the input text prompt.}
    \label{fig:motivation}
\end{figure}

The development of state-of-the-art Vision-Language Foundation models (VLFM), such as Stable Diffusion~\cite{Rombach_2022_CVPR}, has significantly improved the image generation quality and resolution significantly over traditional generative models, such as VAEs and GANs~\cite{gal2022stylegan,brock2018large}. In medical imaging, these new foundation models have demonstrated capability to generate highly realistic 2D images with fine details and textures.
However, diffusion models inherit and amplify data bias~\cite{bansal2022well,perera2023analyzing} from large-scale training data, showing undesired behaviors. For example, when given a text prompt \texttt{A dermatoscopic image with melanoma
showing hairs} to Stable Diffusion fine-tuned on skin cancer data, it generates realistic images with hairs. However, the synthesized images typically also contain well-known artifacts in the dataset, such as a ruler or ink shown in Figure~\ref{fig:motivation}. Thus, a semantic alignment mismatch exists between the text and the synthesized image.

Synthetic image generation is of importance, especially in medical imaging, as it can be used for tasks such as data augmentation~\cite{kebaili2023deep}, debasing classifiers~\cite{kumar2023debiasing} or accurate detection and diagnosis of disease~\cite{goceri2023medical,zhang2017real}. Additionally, high-resolution and precise image-generation capabilities in complex settings, such as drug discovery or personalized diagnosis, require analyzing counterfactual "what if" scenarios~\cite{favero2025conditionaldiffusionmodelsmedical,pawlowski2020deep,kumar2022counterfactual}. 
Recently, Denoising Diffusion models (DDMs)~\cite{ho2020denoising} have shown outstanding performances in high-resolution conditional image generation. Despite their impressive capability to synthesize images, they are prone to biases. These diffusion architectures utilize a controlled sampling process, which can either be classifier-free~\cite{ho2022classifier} or classifier-guided~\cite{dhariwal2021diffusion}. A promising alternative recently proposed for this goal is the use of Reinforcement Learning (RL) to optimize the diffusion process for improved control and adaptability~\cite{black2023training,fan2023dpok,miao2024training,zhang2024large}. Fine-tuning the diffusion model to optimize a desired reward function can enable these models to incorporate task-specific preferences, potentially reducing bias and improving alignment between generated samples and predefined constraints. Denoising Diffusion Policy Optimization (DDPO)~\cite{black2023training} is an RL-based method that reframes the diffusion process as a multi-step Markov Decision Process (MDP) to optimize a given reward function. With the rise of policy-based methods, the effectiveness of various reward functions has been demonstrated in natural imaging domains, including diversity-based rewards~\cite{zhang2024large,miao2024training}, alignment rewards~\cite{black2023training}, and visual rewards such as aesthetic quality~\cite{black2023training}. However their applicability in medical imaging remains underexplored.


In this work, we introduce the first framework for improved performance of text-guided image generation in medical imaging using Stable Diffusion through policy optimization in reinforcement learning. Specifically, we demonstrate that policy-based optimization improves the alignment between the input text prompts and the generated images. We propose a new metric - \textit{Artifact Prevalence Rate (APR)} to compute the presence of the desired attributes in the synthesized image. Extensive experiments are performed on a publicly available dataset, ISIC2019~\cite{tschandl2018ham10000,codella2018skin,combalia2019bcn20000}, which shows that the method synthesizes photorealistic images that are completely in alignment with the medical context.

\section{Methodology}
\begin{figure}[t]
    \centering
    \includegraphics[width=0.9\linewidth]{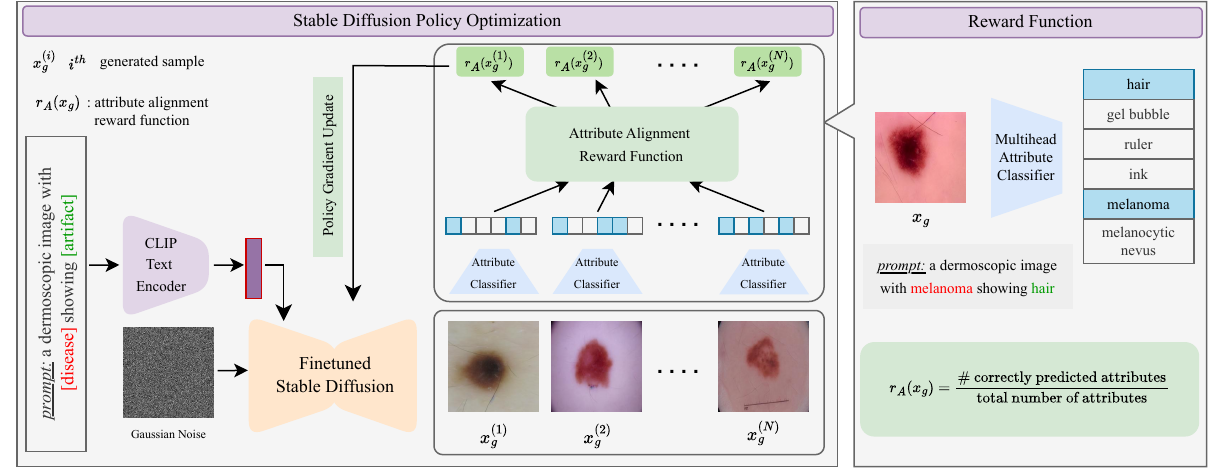}
    \caption{Proposed architecture for policy optimization using a reward function for diverse and realistic image generation using fine-tuned Stable Diffusion. Given an input text prompt, the model synthesizes a realistic image $x_g$ during the reverse diffusion. This generated image is then passed to a pre-trained classifier to compute the reward which helps guide the denoising UNet to improve image synthesis so it is better semantically aligned with the input text.}
    \label{fig:architecture}
\end{figure}

The reinforcement learning framework combines Stable Diffusion in two stages: (i) Fine-tune the original Stable Diffusion v1.5~\cite{Rombach_2022_CVPR} with the medical dataset to align text-image pairs; (ii) Use a pre-trained classifier to compute the reward and update the weights of fine-tuned Stable Diffusion from stage (i), by performing policy optimization. The general framework of our method is now described and illustrated in Figure~\ref{fig:architecture}.

\subsection{Denoising Diffusion Policy Optimization (DDPO)}



\noindent\textbf{Diffusion models:}  Markov chains~\cite{sohl2015deep} model the data generation process by gradually adding and removing noise. The forward process transforms input data \( x_0 \) into Gaussian noise \( x_T \) over \( T \) steps using a variance schedule \( \beta_t \):

\begin{equation*}
q(x_t | x_{t-1}) = \mathcal{N}(\sqrt{1 - \beta_t} x_{t-1}, \beta_t I)
\end{equation*}

Defining \( \alpha_t = 1 - \beta_t \) and \( \bar{\alpha_t} = \prod_{s=1}^{t} \alpha_s \), we can directly express \( x_t \) as:

\begin{equation*}
q(x_t | x_0) = \mathcal{N}(\sqrt{\bar{\alpha_t}} x_0, (1 - \bar{\alpha_t}) I)
\end{equation*}

The reverse diffusion process in DDPMs \cite{ho2020denoising} removes noise using a trained model \( p_\theta(x_{t-1} | x_t) \), approximating the true posterior \( q(x_{t-1} | x_t, x_0) \). The model is trained via variational inference by minimizing the KL divergence between $p$ and $q$.




\noindent\textbf{DDPO:} After the diffusion model is fine-tuned, it can be be further optimized to maximize the expected reward $J(\theta)=\mathbb{E}_{c\sim p(c), x_0\sim p_\theta(x_0|c)}[r(x_0, c)]$ where $p(c)$ is distribution over input text prompts/conditions and $p_\theta(x_0|c)$ is the sample distribution. This is achieved by first re-framing the diffusion model as a multi-step Markov Decision Process (MDP). An MDP can be defined by a set of states S, set of actions A, reward function R, and state transition distribution P. The RL agent tries to maximize the cumulative reward function by learning the policy $\pi(a_t|s_t)$, allowing it to select actions at each step. 

Adapting the framework from \cite{black2023training}, we define the MDP using the denoising model backward process as the policy:

\begin{equation}
\begin{aligned}
    s_t &\triangleq \{x_t, c, t\}, \quad a_t \triangleq x_{t-1} \\
    \pi(a_t | s_t) &\triangleq p_{\theta}(x_{t-1} | x_t, c), \\
    P(s_{t+1} | s_t, a_t) &\triangleq \{\delta(x_{t-1}), \delta(c), \delta(t - 1)\} \\
    \rho(s_{0}) &\triangleq \{\mathcal{N}(0,1), \delta(c), \delta(T)\} \\
    R(s_t, a_t) &\triangleq \mathbbm{1} \{t = 0\} \cdot r(x_0, c),
\end{aligned}
\end{equation}

where the state $s_t$ is defined as the combination of latent $x_t$, time $t$, and the condition $c$. Policy $\pi$ for selecting the action is defined using the denoising model, and $\delta(x)$ denotes the Dirac function. While fine-tuning the diffusion model, we use the importance sampling estimator~\cite{kakade2002approximately} which uses two models $\theta$ and $\theta'$ to calculate the policy gradients for $\theta$: 

\begin{equation}
    \nabla_{\theta} L = \mathbb{E} \left[ \sum_{t=0}^{T} \frac{p_{\theta}(x_{t-1} | x_t, c)}{p_{\theta'}(x_{t-1} | x_t, c)}
    \nabla_{\theta} \log p_{\theta}(x_{t-1} | x_t, c) r(x_0, c) \right].
    \label{eq:policy-gradient}
\end{equation}
The expectation is taken over trajectories (intermediate latents in the denoising process) sampled from the previous model \( \theta' \). This estimator allows multiple gradient evaluations using samples generated from the old model $\theta'$. However, as discussed in \cite{black2023training}, the estimator's accuracy may degrade if \( p_\theta \) and \( p_{\theta'} \) diverge significantly. To address this issue, trust regions \cite{schulman2015trust} are applied to limit the size of the update by using clipping techniques \cite{schulman2017proximal}. This will regularize the change of $\theta$ with respect to $\theta'$ resolving the problem~\cite{miao2024training}.

\subsection{Training Details}
\noindent\textbf{Finetuning Stable Diffusion:} The original Stable Diffusion v1.5~\cite{Rombach_2022_CVPR} architecture consists of four main components: (1) image encoder, (2) Contrastive Language-Image Pre-training (CLIP) text encoder, (3) denoising U-Net for reverse diffusion and (4) an image decoder. Aside from the denoising U-Net, all other components of the Stable diffusion model remain frozen during fine-tuning. Similar to PRISM~\cite{kumar2025prism}, the input to the CLIP encoder is a template text: \texttt{A dermoscopic image with [disease] showing [artifact]}, where \texttt{disease} is melanoma or melanocytic nevis and \texttt{artifact} is hairs, gel bubbles, ink or ruler thus helping to synthesize semantically aligned images from the input text.

\noindent\textbf{Policy based reward:} An $H$-head Efficient-Net~\cite{tan2019efficientnet} classifier ($H=6$) is trained to identify the presence of \texttt{artifacts} including melanoma and melanocytic nevus. The reward function, $r_A(.)$, with respect to a generated image $x_g$ is computed as a ratio of the number of attributes correctly predicted by the pre-trained Efficient-Net classifier to the total number of attributes. This reward function is then used to update the weights of denoising U-Net. This change in the value of the reward function between successive iterations is used as a stopping criterion for model fine-tuning. This is discussed in Algorithm~\ref{alg:rl-finetune}.

\begin{algorithm}[t]
\caption{Stable Diffusion Models with RL implementing policy based gradient update.}
\label{alg:rl-finetune}
\begin{algorithmic}[1]
\Require Pre-trained diffusion model $p_{\theta_{pre}}$, attribute classifier $f(\mathbf{X})$, reward function $r_A(c_{pred}, c)$.
\State Initialize $p_{\theta} = p_{\theta_{pre}}$
\While{$\theta$ not converged}
    \For{each prompt $c \sim p(c)$}
        \State Sample $M$ images $\mathbf{X}_g | c = \{ x_{0,g}^1, \dots, x_{0,g}^M \}, x_{0,g}^m \sim p_{\theta}(x_0 | c)$, together with their intermediate states $x_{1:T,g}^m$.
        \State Compute individual rewards with the attribute reward function for each condition $c$, $r_A(f(\mathbf{X}_g); c)$.
        \State Take $K$ rounds of policy gradient steps with \ref{eq:policy-gradient}.
    \EndFor
\EndWhile
\State \Return Fine-tuned diffusion model $p_{\theta}$.
\end{algorithmic}
\end{algorithm}

\subsection{Metrics \& Evaluation of Synthesized Samples}
The synthesized samples from our method, SD+RL, are compared against the fine-tuned Stable Diffusion (baseline), SD. Both these methods generate realistic samples, but the synthesized images aren't always perfectly aligned with the input prompt. Thus, to measure the semantic alignment between image-text pairs, we propose a new metric, \textit{Artifact Prevalence Rate (APR)}, computed as follows:
\begin{equation}\label{eq:apr}\text{APR} = \frac{\text{count of } x_i \in X: f(x_i) = C(\text{input text})  }{N}
\end{equation}
$X$ are all the synthesized samples for the attribute under observation, $N$ is the number of synthesized samples, $f(.)$ is the multi-head attribute classifier that identifies hair, gel bubbles, rule and ink, $C(.)$ is the function that one-hot encodes the \texttt{disease} and \texttt{artifact} information in the input text. A higher value of APR is expected as it would indicate that only the attribute mentioned in the text is prevalent in the synthesized samples and all others are ignored.

Finally, if the synthesized samples carry rich discriminative information about the domains, they should be able to improve the performance of subclasses with fewer real samples through augmentation. As such, the dataset was augmented with synthesized samples and the classifier's performance per subclass was evaluated before and after augmentation.

\section{Experiments and Results}
\subsection{Dataset and Implementation Details}
We perform experiments on a publicly available dataset, ISIC 2019~\cite{tschandl2018ham10000,codella2018skin,combalia2019bcn20000}. The parameters and code to fine-tune the Stable Diffusion model are made publicly available\footnote {The Github repository will be made public upon acceptance}. Table~\ref{table:samples} shows the distribution of samples and the artifacts as per train, validation and test splits.

\begin{table}[t]
\caption{Summary of train, validation, and test splits for ISIC 2019 dataset. Note that the prevalence of ink is significantly low compared to others.}
\centering
\begin{tabular}{@{}lcccccc@{}}
\toprule
 & Melanoma & \begin{tabular}[c]{@{}c@{}}Melanocytic\\Nevus\end{tabular} & Hair & \begin{tabular}[c]{@{}c@{}}Gel\\Bubbles\end{tabular} & Ink & Ruler \\
\midrule
Train & 2750 & 9254 & 4514 & 1300 & 201 & 1608 \\
Validation & 454 & 1665 & 802 & 228 & 34 & 308 \\
Test & 537 & 1956 & 989 & 257 & 37 & 341 \\
\bottomrule
\end{tabular}
\label{table:samples}
\end{table}

To evaluate and compare our method (RL+SD) with the baseline (SD), we use Fréchet Inception Distance (FID)~\cite{heusel2017gans} and LPIPS~\cite{zhang2018unreasonable}, in addition to the proposed APR metric. For a comprehensive analysis, we generate approximately 70K samples across various prompts and label combinations, ensuring  robust evaluation of both models.

\subsection{Qualitative Evaluations} 
\begin{figure}[t]
    \centering
    \includegraphics[width=0.8\linewidth]{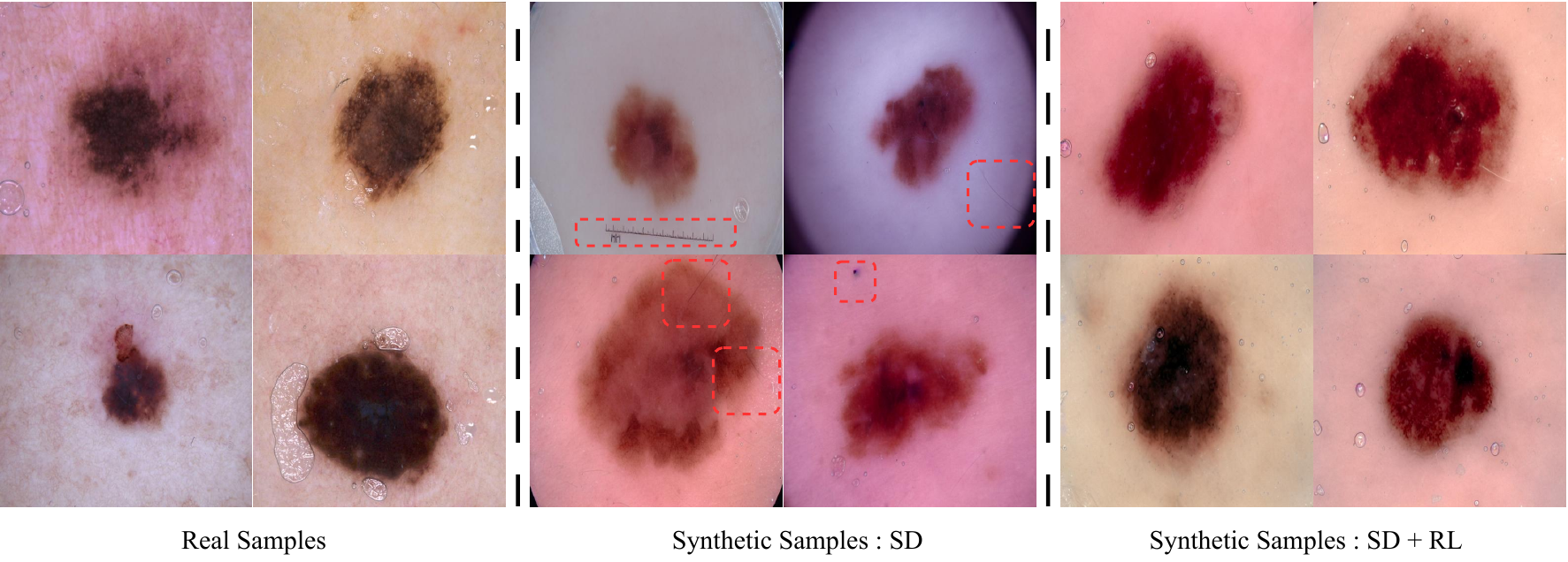}
    \caption{Comparing real samples for the category "melanocytic nevus with gel bubbles" with the synthesized images using fine-tuned Stable Diffusion (SD) and fine-tuned Stable Diffusion with reinforcement learning (SD+RL). Note  the unwanted \textcolor{red}{artifacts} present in the image synthesized by SD.}
    \label{fig:qual}
\end{figure}

Melanomas often have irregular, asymmetric shapes, while benign melanocytic nevus are typically well-circumscribed and symmetric~\cite{damsky2017melanocytic,elder2006precursors}. Additionally, melanomas tend to have irregular, poorly defined borders and are often large with a diameter of about 6mm, while melanocytic nevus have clear, well-defined edges~\cite{sung2022nevi,elder2006precursors}. In Figure~\ref{fig:qual}, both the methods can accurately replicate the distinct melanomas and melanocytic nevus characteristics in the generated images. However, the samples from the method using SD+RL perform better as they create fewer or no unwanted attributes while synthesizing the images. 

\begin{figure}[t]
    \centering
    \includegraphics[width=0.8\linewidth]{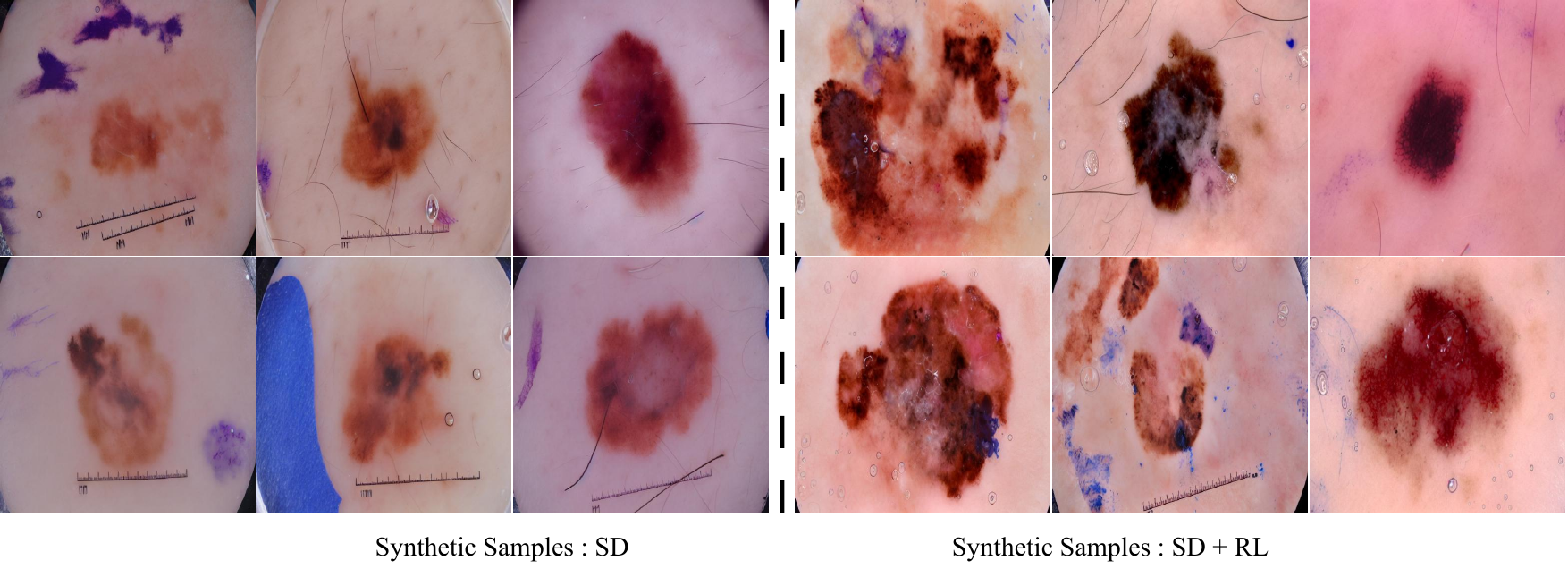}
    \caption{Qualitative comparisons of synthesized images of subgroups (based on combinations of disease and artifacts) for which none or a few (less than 20) real samples are present. Note that some of these subgroups include combinations of attributes, such as melanoma with gel bubbles and ink or melanocytic nevus with ink and hair. }
    \label{fig:qual2}
\end{figure}

In Figure~\ref{fig:qual2}, we show that the proposed framework can create better samples for domains with no corresponding real samples i.e., during fine-tuning the Stable Diffusion model never saw these combinations of artifacts. For example, no real samples of melanocytic nevus with attributes hairs and ink are in the training data and SD+RL is able to synthesize visually better samples with better alignment and higher artifact quality. 

\subsection{Quantitative Evaluations}
Table~\ref{tab:apr} evaluates the model using the APR metric across different disease and artifact combinations. The ability of our method (SD+RL) to maintain high APR values across underrepresented classes further validates the effectiveness of our optimization strategy. In Table~\ref{tab:sd-vs-rl}, our method outperforms the pre-trained Stable Diffusion (SD) model with respect to APR, indicating that our method synthesizes images better aligned with the input prompts and generates fewer unwanted artifacts. Additionally, our methods also achieve a lower FID and similar LPIPS compared to the baseline. Finally, Table~\ref{tab:augment} presents the performance of the model after augmenting the training dataset with synthesized images. The results show that classifiers trained on real data augmented with our model's generated images achieve the highest performance. This confirms that our synthesized samples provide useful augmentations that enhance classifier accuracy, further validating the effectiveness of our approach in improving downstream tasks.

\begin{table}[t]
\centering
\caption{Quantitative results for evaluating synthesized samples (1000 samples per class). Note the subgroups MEL with hair, ink and NV with ink have fewer real samples. Also, SD+RL consistently achieves higher APR values, even for classes with a limited number of real samples. (\textbf{Gel} = Gel Bubbles) }

{\fontsize{8}{9.4}\selectfont
\begin{tabular}{llcccc|cccc}
\toprule
 &  & \multicolumn{4}{c|}{\textbf{Melanoma (MEL)}} & \multicolumn{4}{c}{\textbf{Melanocytic Nevus (NV)}} \\ 
\cmidrule(lr){3-6} \cmidrule(lr){7-10}
 &  & \textbf{Hair} & \textbf{Gel} & \textbf{Ink} & \textbf{Ruler} & \textbf{Hair} & \textbf{Gel} & \textbf{Ink} & \textbf{Ruler} \\ 
\midrule
\# Real Sample &  & 543 & 260 & 4 & 440 & 3051 & 489 & 21 & 520 \\ 
\midrule
\multirow{2}{*}{APR $\uparrow$ (\%)} 
& SD (Baseline)   &  63.74 & 13.75  & 0.2  & 76.56  & 44.37  & 13.54  & 0.1  & 80  \\
& \textbf{SD+RL} & \textbf{86.97} & \textbf{94.37} & \textbf{1.6} & \textbf{93.85} & \textbf{75.72} & \textbf{80.62} & \textbf{1.4} & \textbf{87.18} \\  

\bottomrule
\end{tabular}}
\label{tab:apr}
\end{table}

\begin{table}[htbp]
\centering

\caption{Quantitative results for evaluating synthesized samples for 32 different prompts (1000 samples per prompt) of various artifacts with MEL or NV. LPIPS was calculated by randomly selecting 1000 samples per prompt and comparing them with corresponding real images, resulting in a total of 32,000 comparisons.}
\begin{tabular}{l c c c}
\toprule
\textbf{Model} & \textbf{APR$\uparrow$(\%)} & \textbf{FID$\downarrow$} & \textbf{LPIPS$\downarrow$} \\ 
\midrule
SD (Baseline)    & 18.28  & 121.47  & 0.60 \\
\textbf{SD+RL} & \textbf{63.14} & \textbf{114.7} & \textbf{0.59} \\ 
\bottomrule
\end{tabular}
\label{tab:sd-vs-rl}
\end{table}

\begin{table}[htbp]
\centering
\caption{F1 and Accuracy for different attributes on ISIC test set for Real, RL-synthesized+Real, and SD-synthesized+Real. The metrics are calculated for each attribute independently of the others.}
\label{tab:mytable}
\begin{tabular}{l cc cc cc}
\toprule
 & \multicolumn{2}{c}{Real}
 & \multicolumn{2}{c}{\textbf{SD-RL+Real}}
 & \multicolumn{2}{c}{SD+Real}\\
\cmidrule(lr){2-3}\cmidrule(lr){4-5}\cmidrule(lr){6-7}
\textbf{Setting}
& \textbf{F1} & \textbf{Accuracy}
& \textbf{F1} & \textbf{Accuracy}
& \textbf{F1} & \textbf{Accuracy} \\
\midrule
Hair                           & 0.91 & 93.10 & \textbf{0.92} & \textbf{93.54} & 0.91 & 93.26 \\
Gel Bubbles                    & 0.66 & 93.90 & \textbf{0.70 }& \textbf{94.06} & 0.67 & 93.90 \\
Ruler                          & 0.89 & 96.95 & \textbf{0.89 }& \textbf{97.03} & 0.85 & 96.23 \\
Ink                            & \textbf{0.90} & \textbf{99.72} & 0.89 & 99.68 & 0.86 & 99.60 \\
\bottomrule
\end{tabular}
\label{tab:augment}
\end{table}

\section{Conclusion}
In this work, we demonstrate the first method showing alignment between the text prompt and image generation using a vision-foundation model guided by a policy optimization for medical imaging applications. We show through extensive qualitative and quantitative validation that these images align well with the input text prompt, and they are helpful for downstream tasks such as augmenting the classifier to improve performance over minority classes. Future work will explore the use of diverse policies for complex tasks such as subgroup clustering in the latent space for disease image marker discovery.

\section*{Acknowledgments}
The authors are grateful for funding provided by the Natural Sciences and Engineering Research Council of Canada, the Canadian Institute for Advanced Research (CIFAR) Artificial Intelligence Chairs program, Mila - Quebec AI Institute, Google Research, Calcul Quebec, and the Digital Research Alliance of Canada.
\bibliographystyle{splncs04}
\bibliography{main}

\end{document}